\documentclass[conference]{IEEEtran}
\IEEEoverridecommandlockouts

\usepackage{cite}
\usepackage{amsmath,amssymb,amsfonts}
\usepackage{algorithmic}
\usepackage{graphicx}
\usepackage{textcomp}
\usepackage{xcolor}
\def\BibTeX{{\rm B\kern-.05em{\sc i\kern-.025em b}\kern-.08em
    T\kern-.1667em\lower.7ex\hbox{E}\kern-.125emX}}
\begin{document}

\title{CAKD: A Correlation-Aware Knowledge Distillation Framework Based on Decoupling Kullback-Leibler Divergence\\
\thanks{The first author also would like to acknowledge the scholarship supported by China Scholarship Council.}
}

\author{\IEEEauthorblockN{Zao Zhang\textsuperscript{1,2}, Huaming Chen\textsuperscript{2}, Pei Ning\textsuperscript{3}, Nan Yang\textsuperscript{2}, Dong Yuan\textsuperscript{2}}
\IEEEauthorblockA{
\textsuperscript{1}Institute of Automation, Chinese Academy of Sciences, Beijing, China\\
\textsuperscript{2}Faculty of Engineering, The University of Sydney, Sydney, Australia\\
\textsuperscript{3}School of Mechanical and Electronic Engineering, Wuhan University of Technology, Wuhan, China\\
zao.zhang@ia.ac.cn, huaming.chen@sydney.edu.au, nan.yang@sydney.edu.au\\ pei.ning@whut.edu.cn, dong.yuan@sydney.edu.au}
}



\maketitle

\begin{abstract}
In knowledge distillation, a primary focus has been on transforming and balancing multiple distillation components. In this work, we emphasize the importance of thoroughly examining each distillation component, as we observe that not all elements are equally crucial. From this perspective, we decouple the Kullback-Leibler (KL) divergence into three unique elements: Binary Classification Divergence (BCD), Strong Correlation Divergence (SCD), and Weak Correlation Divergence (WCD). Each of these elements presents varying degrees of influence. Leveraging these insights, we present the Correlation-Aware Knowledge Distillation (CAKD) framework. CAKD is designed to prioritize the facets of the distillation components that have the most substantial influence on predictions, thereby optimizing knowledge transfer from teacher to student models. Our experiments demonstrate that adjusting the effect of each element enhances the effectiveness of knowledge transformation. Furthermore, evidence shows that our novel CAKD framework consistently outperforms the baseline across diverse models and datasets. Our work further highlights the importance and effectiveness of closely examining the impact of different parts of distillation process.
\end{abstract}


\section{Introduction}
Knowledge distillation (KD) operates on a teacher-student paradigm, where a streamlined and efficient student model is trained to replicate the operations of a larger, more sophisticated teacher model. Current methodologies frequently treat the distillation component as a singular unit, potentially neglecting the nuanced effects that its individual elements might have on the model's predictions. This perspective could lead to an incomplete understanding of how different aspects of the distillation process contribute to the overall performance and efficiency of the student model. By delving deeper into the intricacies of the distillation component, we aim to uncover the underlying impacts that its various elements can have, thereby enhancing the effectiveness of the KD process and improving the resulting student models.

In this work, we leverage the insights gleaned from model pruning methods, which have demonstrated that not all intermediate features or logits carry the same weights in determining a model's final output. Some features exert a significant influence on the model's decision-making process, while others may be less impactful or even redundant for the specific task at hand. By applying similar principles to knowledge distillation, we aim to identify and emphasize the most critical elements of the distillation component, thereby enhancing the efficiency and effectiveness of the student model.

To achieve the aforementioned objectives, we selected the KL divergence, a commonly used loss function for knowledge distillation, as the primary focus of our study. The rationale for this choice lies in the fact that, through mathematical decoupling, the KL divergence can be decomposed losslessly into three sub-KL divergences. This decomposition allows us to attribute the correlations between important features (or logits), unimportant features (or logits), and the interactions between important and unimportant features (or logits) to distinct sub-KL divergences, thereby providing them with tangible physical interpretations. We have designated the three sub-KL divergence formulas resulting from this assignment as Strong Correlation Divergence (SCD), Weak Correlation Divergence (WCD), and Binary Classification Divergence (BCD), respectively. By rationally adjusting the effect of SCD, WCD, and BCD on the distillation process, we can help student models selectively learn high-quality information and improve their final performance. 

Our main contributions can be distilled as follows:

\begin{itemize}
    \item We offer a fresh perspective, delving into the effect of individual elements within the distillation component by decoupling the KL divergence.
    \item Through empirical analysis, we demonstrate that enhancing the impacts of SCD and WCD allows for tailored influence over their respective roles in improving the student model's performance.
    \item We introduce the framework CAKD, which delivers superior performance compared to the state-of-the-art methods across a wide range of models and datasets.
\end{itemize}

\section{Related Works}
\label{sec:realted}

Recent development of KD focus on two primary categories: logit-based KD and feature-based KD. Logit-based KD focuses on the student model's emulation of the teacher model's output logits. This method utilizes both traditional `hard' labels and nuanced `soft' labels generated by the teacher model during training. The soft labels provide rich information about the relationships between classes, offering a comprehensive explanation of the output class distribution. By aligning its outputs with these soft labels, the student model inherits the knowledge encapsulated in the teacher model's outputs. Notable works in this area include studies by \cite{kim2021comparing}, and \cite{ding2021knowledge}, which have demonstrated the efficacy of this approach in various applications. In contrast, feature-based KD aims to go beyond merely matching the final outputs. This approach seeks to align the student model’s intermediate feature representations or activations with those of the teacher model. The underlying idea is that these intermediate features, similar to the stages of phased learning in humans, carry essential insights that help the student model develop a deeper understanding of the data. By encouraging the student model to mimic the teacher's internal thought process, feature-based KD enhances the learning experience and improves the student model's performance. Vaious strategies have been explored to optimize this alignment \cite{chen2021distilling, xu2020kernel}. These advancements have made KD a more versatile and powerful tool for model compression, capable of producing student models that are both efficient and high-performing. CAKD is designed to work independently with either logits or features, or to merge both methods, depending on the requirements of the task at hand. By allowing for the combination of logit-based and feature-based techniques, CAKD leverages the strengths of both approaches, ensuring that the student model benefits from comprehensive guidance. 


\begin{figure*}
  \centering
  \includegraphics[width=0.8\textwidth]{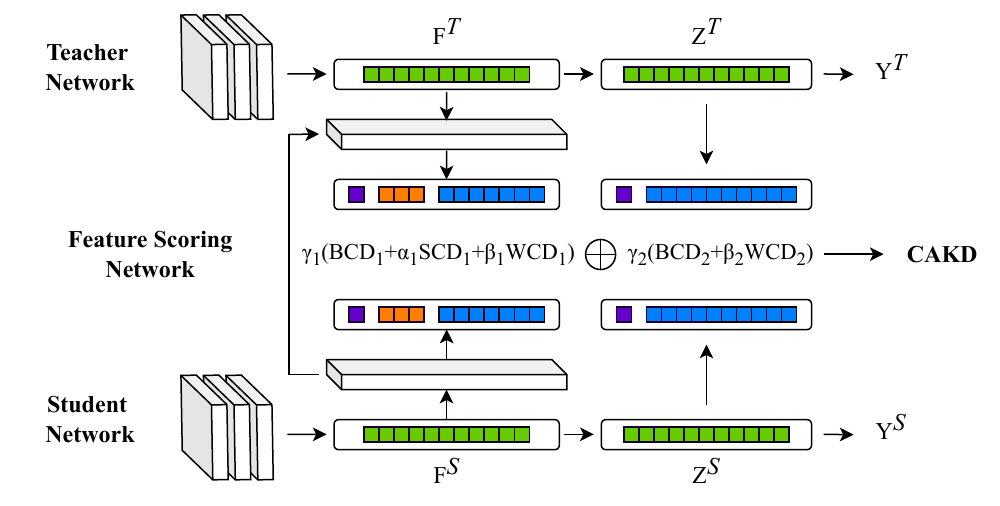}
  \caption{Overview of the Correlation-Aware Knowledge Distillation(CAKD).}
\label{fig:overview}
\end{figure*}

\section{Correlation-Aware Knowledge Distillation}
\label{sec:method}
In this section, we propose and formalize the KL-Divergence decoupling mechanism and knowledge distillation process.

\subsection{Overview}
\label{sec:overview}

The architecture of our proposed CAKD is illustrated in Fig.\ref{fig:overview}. In this figure, the teacher network's features and logits are represented as $F^T$ and $Z^T$, while the student network's counterparts are marked as $F^S$ and $Z^S$. These elements are visualized as green squares. Following this, these elements are classified into three categories: elements strongly correlated to the prediction, elements weakly correlated to the prediction, and the precision associated with this binary classification of correlation. These categories are depicted as orange, blue, and purple squares, respectively.

The CAKD loss, our global loss function, encompasses differences founded on KL divergence across features and logits. Each component is further decoupled into a weighted sum of multiple KL divergences. When we delve into intermediate features, the SCD, WCD, and BCD respectively represent the KL-Divergence in strong correlation clusters between the teacher and student networks, weak correlation clusters between the teacher and student networks, and the delineation of strong and weak correlations between the teacher and student networks. The intricacies of this decoupling process are detailed in Sec.\ref{sec:rkl}.

For the logit-based component, only WCD and BCD come into play, capturing the KL divergence in weak correlation embeddings and the KL divergence in the delineation of strong and weak correlations between the teacher network and student network, respectively. The rationale for only two categories here stems from the scenario when our task has just one ground truth label. In such cases, the differentiation of SCD is encompassed into BCD. Similar to the feature-based component, the decoupling detail is introduced in Sec.\ref{sec:rkl} as well.

CAKD furnishes a robust methodology to scrutinize the influence of individual elements within each distillation component. Based on our experimental findings, when operating with a highly confident teacher model, the correlation-based classification is more straightforward, indicating a reduced BCD value. In such contexts, both SCD and WCD, which transmit intricate knowledge, are subdued within standard KL divergence. The equilibrium between SCD and WCD is intrinsically tied to the teacher model's confidence. For instance, when the teacher model exudes high confidence, SCD gains prominence. Conversely, with a teacher model exhibiting lower confidence, pinpointing highly correlated features or logits becomes challenging, amplifying WCD's role.

CAKD provides us with a good way to study the impact of each element within each distillation component. In traditional KL divergence, SCD and WCD will be suppressed, the reason is explored by the mathematical derivation in Sec.\ref{sec:rkl}. Our proposed CAKD achieves a great effect on enhancing the SCD and WCD. Further, according to our experiments,  SCD and WCD transfer difficult knowledge which has a high effect on the accuracy. The effect of SCD and WCD is highly related to the confidence of the teacher model. 


\subsection{Decouple Kullback-Leibler Divergence}
\label{sec:rkl}
KL divergence measures the differences between two probability distributions. In KD, KL divergence is usually adopted to measure the difference between the teacher model and the student model, which can be written as:
\begin{align}
\label{eq:1}
    \begin{aligned}
    \mathrm{KD} & =\mathrm{KL}\left(\mathbf{p}^{\mathcal{T}} \| \mathbf{p}^{\mathcal{S}}\right) \\
    & =\sum_{i=1}^C p_i^{\mathcal{T}} \log \left(\frac{p_i^{\mathcal{T}}}{p_i^{\mathcal{S}}}\right) \\
    \end{aligned}
\end{align}

The probability can be denoted as $\mathbf{p}=\left[p_1, p_2, \ldots, p_t, \ldots, p_C\right] \in \mathbb{R}^{1 \times C}$, where $p_i$ is the probability of the $i$-th feature and $C$ is the number of features. Each element in $p$ can be obtained by the softmax function:

\begin{equation}
\label{eq:2}
    p_i=\frac{\exp \left(f_i\right)}{\sum_{i=1}^C \exp \left(f_i\right)}
\end{equation}

The success of filter pruning has proven that in a feature map, many features are not that important to the prediction, indicating they can be pruned. Based on this concept, we categorize features into two clusters: one cluster strongly correlates with the prediction which is represented by $\mathbf{S}$, the number of features in cluster $\mathbf{S}$ is $C_s$, and another cluster contains features weakly correlates with the prediction which is represented by $\mathbf{W}$, the number of features in cluster $\mathbf{W}$ is $C_w$. Then we define the following notations,  $\mathbf{b}=\left[p_s,p_w\right] \in \mathbb{R}^{1 \times 2}$ represents the binary probabilities of classification between the strong correlation cluster ($p_s$) and the weak correlation cluster ($p_w$), which can be calculated by:

\begin{equation}
\begin{aligned}
\label{eq:3}
    p_s&=\frac{\sum_{i=1, i \in S}^C \exp \left(f_i\right)}{\sum_{k=1}^C \exp \left(f_k\right)}  \\
    p_w&=\frac{\sum_{i=1, i \notin S}^C \exp \left(f_i\right)}{\sum_{k=1}^C \exp \left(f_k\right)}
\end{aligned}
\end{equation}

Then, we declare $\hat{\mathbf{p}}=\left[\hat{p}_1, \hat{p}_2, \hat{p}_t, \ldots, \hat{p}_C\right] \in \mathbb{R}^{1 \times C}$ to independently model probabilities among features. Correspondingly, we declare $\hat{\mathbf{p_s}} \in \mathbb{R}^{1 \times C_s}$ and $\hat{\mathbf{p_w}} \in \mathbb{R}^{1 \times C_w}$ to model the probabilities among features with strong correlations and weak correlations. Each element in $\hat{\mathbf{p_s}}$ or $\hat{\mathbf{p_w}}$  is calculated by  
\begin{equation}
\label{eq:4}
    \hat{p_i}=
    \begin{cases}
    \frac{\exp \left(f_i\right)}{\sum_{i=1,i \in S}^C \exp \left(f_i\right)} & \text{if } i \in S \\
    \\
    \frac{\exp \left(f_i\right)}{\sum_{i=1,i \notin S}^C \exp \left(f_i\right)} & \text{if } i \notin S
    \end{cases}
\end{equation}   

Based on Eqn.\eqref{eq:2}, Eqn.\eqref{eq:3} and Eqn.\eqref{eq:4}, we get:
\begin{equation}
\label{eq:5}
    p_i = 
    \begin{cases} 
    p_s \cdot \hat{p_i}  & \text{if } i \in S \\
    p_w \cdot \hat{p_i}  & \text{if } i \notin S 
    \end{cases}
\end{equation}

Based on the above equations, we can rewrite the Eqn.\eqref{eq:1} as:
\begin{equation}
\label{eq:6}
    \begin{aligned}
    \mathrm{KD} 
    & =\sum_{i \in S}  p_s^{\mathcal{T}} \hat{p_i}^{\mathcal{T}} \log \left(\frac{p_ s^{\mathcal{T}} \hat{p_i}^{\mathcal{T}}}{p_s ^{\mathcal{S}} \hat{p_i}^{\mathcal{S}}}\right) +\sum_{i \notin S}  p_w^{\mathcal{T}} \hat{p_i}^{\mathcal{T}} \log \left(\frac{p_w^{\mathcal{T}} \hat{p_i}^{\mathcal{T}}}{p_w^{\mathcal{S}} \hat{p_i}^{\mathcal{S}}}\right) \\
    & =p_s^{\mathcal{T}} \sum_{i \in  S}  \hat{p_i}^{\mathcal{T}}\left(\log \left(\frac{\hat{p_i}^{\mathcal{T}}}{\hat{p_i}^S}\right)+\log \left(\frac{p_s^{\mathcal{T}}}{p_s^{\mathcal{S}}}\right)\right) \\
    & \qquad  +p_w^{\mathcal{T}} \sum_{i \notin S}  \hat{p_i}^{\mathcal{T}}\left(\log \left(\frac{\hat{p_i}^{\mathcal{T}}}{\hat{p_i}^S}\right)+\log \left(\frac{p_w^{\mathcal{T}}}{p_w^{\mathcal{S}}}\right)\right) \\
    & = \underbrace{p_s^{\mathcal{T}}  \log \left(\frac{p_ s^{\mathcal{T}}}{p_s^{\mathcal{S}}}\right)
    +p_w^{\mathcal{T}} \log \left(\frac{p_w^{\mathcal{T}}}{p_w^{\mathcal{S}}}\right)}_{\mathrm{KL}\left(\mathbf{b}^{\mathcal{T}} \| \mathbf{b}^{\mathcal{S}}\right)} \\
    & \qquad + \underbrace{p_s^{\mathcal{T}}\sum_{i \in S}  \hat{p_i}^{\mathcal{T}} \log \left(\frac{\hat{p_i}^{\mathcal{T}}}{\hat{p_i}^{\mathcal{S}}}\right)}_{p_s^{\mathcal{T}}\mathrm{KL}\left(\mathbf{\hat{p_s}}^{\mathcal{T}} \| \mathbf{\hat{p_s}}^{\mathcal{S}}\right)} \\
    & \qquad + \underbrace{p_w^{\mathcal{T}} \sum_{i \notin S}\hat{p_i}^{\mathcal{T}} \log \left(\frac{\hat{p_i}^{\mathcal{T}}}
    {\hat{p_i}^{\mathcal{S}}}\right)}_{p_w^{\mathcal{T}} \mathrm{KL}\left(\mathbf{\hat{p_w}}^{\mathcal{T}} \|\mathbf{\hat{p_w}}^{\mathcal{S}}\right)}\\
    \end{aligned}
\end{equation}

By Eqn.\eqref{eq:6}, we decoupled the KL divergence into a weighted sum of three independent KL divergence, where the three parts respectively measure the accuracy of the binary classification between $\mathbf{S}$ and $\mathbf{W}$; the divergence in feature distributions between the teacher model and the student model within $\mathbf{S}$, and the divergence in feature distributions between the teacher model and the student model within $\mathbf{W}$. To discuss the effect of the three parts more easily, the Eqn.\eqref{eq:6} can be abbreviated as:
\begin{equation}
\label{eq:7}
\begin{aligned}
\mathrm{KD} &=\mathrm{KL}\left(\mathbf{b}^{\mathcal{T}} \| \mathbf{b}^{\mathcal{S}}\right) \\
 & +p_s^{\mathcal{T}} \mathrm{KL}\left(\mathbf{\hat{p_s}}^{\mathcal{T}} \| \mathbf{\hat{p_s}}^{\mathcal{S}}\right)+p_w^{\mathcal{T}} \mathrm{KL}\left(\mathbf{\hat{p_w}}^{\mathcal{T}} \| \mathbf{\hat{p_w}}^{\mathcal{S}}\right) \\ &=\mathrm{BCD}+p_s^{\mathcal{T}}\mathrm{SCD}+p_w^{\mathcal{T}} \mathrm{WCD}
\end{aligned}
\end{equation}

\begin{table*}[t]
\caption{Top-1 accuracies(100\%) on the CIFAR-100.} 
\begin{center}
\begin{tabular}{c|ccccc}
  \hline
    & \multicolumn{3}{c}{Same Architecture Style} & \multicolumn{2}{c}{Different Architecture Style} \\
  \hline
  Teacher  & ResNet56 & ResNet32x4& WRN40-2  & ResNet32x4 & WRN40-2 \\
  Acc   &   72.32   &  79.32 &  75.50  &  79.32   &  75.50   \\
  Student & ResNet20 & ResNet8x4  & WRN16-2& ShuffleNetV2 & ShuffleNetV1 \\
  Acc &   69.01   &  72.24  &  72.93  & 71.42 & 70.17 \\
  \hline
  KD\cite{hinton2015distilling}&70.38&74.59&74.58&74.51&74.81\\
  CRD\cite{tian2019contrastive}&70.89&75.03&75.43&75.66&75.94\\
  OFD\cite{heo2019comprehensive}&70.68&74.80&75.19&76.84&75.79\\
 CTKD\cite{li2023curriculum}&70.98&75.07&75.16&75.92&75.77\\
 ReviewKD\cite{chen2021distilling}&71.67&75.43&76.01&77.75&76.98\\
  DKD\cite{zhao2022decoupled}&71.69&75.72&75.63&77.25&76.59\\
 NKD\cite{yang2023knowledge}&71.98&76.02&75.74&77.61&76.83\\
  \hline
  CAKD(logit \& single feature)&\textbf{72.27}&\textbf{76.11}&\textbf{76.04}&\textbf{77.94}&\textbf{77.12}\\
  \hline

\end{tabular}
\label{tab:cifar100}
\end{center}
\end{table*}

\begin{table*}[t]
\caption{Top-1 accuracies(100\%) on the Tiny-ImageNet.} 
\begin{center}
\begin{tabular}{c|ccccc}
  \hline
   & \multicolumn{3}{c}{Same Architecture Style} & \multicolumn{2}{c}{Different Architecture Style} \\
  \hline
  Teacher  & ResNet56 & ResNet32x4 & WRN40-2 & ResNet32x4 & WRN40-2 \\
  Acc   &   50.78   &  60.17  &   56.06 & 60.17     &  56.06   \\
  Student & ResNet20 & ResNet8x4& WRN16-2 & ShuffleNetV2  & ShuffleNetV1 \\
  Acc &   42.66   &  47.2   & 48.71 &  47.04 & 46.1 \\
  \hline
  KD\cite{hinton2015distilling}&43.61&50.57&50.21&51.46&49.13\\
  CRD\cite{tian2019contrastive}&43.96&51.29&50.60&52.18&49.92\\
  OFD\cite{heo2019comprehensive}&43.85&51.22&50.39&52.97&49.79\\
  CTKD\cite{li2023curriculum}&44.01&51.26&50.41&52.41&49.68\\
  ReviewKD\cite{chen2021distilling}&44.38&51.38&51.05&53.61&50.65\\
  DKD\cite{zhao2022decoupled}&44.44&51.51&51.10&53.33&50.36\\
  NKD\cite{yang2023knowledge}&44.67&51.48&51.24&53.85&50.61\\
  \hline
  CAKD(logit \& single feature)&\textbf{44.93}&\textbf{51.66}&\textbf{51.43}&\textbf{54.66}&\textbf{50.97}\\
  \hline
\end{tabular}
\label{tab:tiny}
\end{center}
\end{table*}

For tasks with multiple ground truth labels, the logit works in the same way as features with Eqn.\eqref{eq:7}. For the  task with only one ground truth label, if our target embedding for distillation is the logits rather than the intermediate features, the strong correlation cluster will contain only one element corresponding to the ground truth label, which can be rewritten as:

\begin{equation}
\label{eq:8}
    \begin{aligned}
    \begin{aligned}
    \mathrm{KD} & =p_s^{\mathcal{T}} \log \left(\frac{p_s^{\mathcal{T}}}{p_s^{\mathcal{S}}}\right)+p_w^{\mathcal{T}} \sum_{i \notin S}^C \hat{p}_i^{\mathcal{T}}\left(\log \left(\frac{\hat{p}_i^{\mathcal{T}}}{\hat{p}_i^{\mathcal{S}}}\right)+\log \left(\frac{p_w^{\mathcal{T}}}{p_w^{\mathcal{S}}}\right)\right) \\
    & =\underbrace{p_s^{\mathcal{T}} \log \left(\frac{p_s^{\mathcal{T}}}{p_s^{\mathcal{S}}}\right)+p_w^{\mathcal{T}} \log \left(\frac{p_w^{\mathcal{T}}}{p_w^{\mathcal{S}}}\right)}_{\mathrm{KL}\left(\mathbf{b}^{\mathcal{T}} \| \mathbf{b}^{\mathcal{S}}\right)}+\underbrace{p_w^{\mathcal{T}} \sum_{i \notin S}^C \hat{p}_i^{\mathcal{T}} \log \left(\frac{\hat{p}_i^{\mathcal{T}}}{\hat{p}_i^{\mathcal{S}}}\right)}_{p_w^{\mathcal{T}}\mathrm{KL}\left(\hat{\mathbf{p}}^{\mathcal{T}} \| \hat{\mathbf{p}}^{\mathcal{S}}\right)}
\end{aligned}
    \end{aligned}
\end{equation}

Finally, the Eqn.\eqref{eq:6} will be reformulated as:
\begin{equation}
\label{eq:9}
\begin{aligned}
 \mathrm{KD} &=\mathrm{KL}\left(\mathbf{b}^{\mathcal{T}} \| \mathbf{b}^{\mathcal{S}}\right)+p_w^{\mathcal{T}} \mathrm{KL}\left(\mathbf{\hat{p_w}}^{\mathcal{T}} \| \mathbf{\hat{p_w}}^{\mathcal{S}}\right) \\ &=\mathrm{BCD}+p_w^{\mathcal{T}} \mathrm{WCD}
\end{aligned}
\end{equation}

Through the aforementioned reformulation, we have successfully decoupled the KL divergence into a weighted sum of multiple KL divergences. This facilitates independent analysis of each element and enables us to adjust their respective weights to get better performance.

\section{Experiments}
\label{sec:experi}
In this section, we empirically investigate the effectiveness of the proposed CAKD framework. The experimental setup is detailed in Sec. \ref{sec:setup}, where we outline the datasets, models, and evaluation metrics used in our experiments. The main results are presented in Sec. \ref{sec:result}, showcasing the performance of CAKD across various benchmarks. These studies provide insights into the contributions of individual elements to the overall performance of the model, demonstrating the robustness and effectiveness of our proposed approach.

\subsection{Experimental Setup}
\label{sec:setup}
\subsubsection{Datasets}
\label{sec:dataset}
We evaluate our method using the (1) Cifar-100 \cite{krizhevsky2009} dataset, which contains 50,000 training images of 100 classes. Each class has 500 training images and 100 test images. (2) Tiny-ImageNet \cite{tiny-imagenet} dataset is a subset of ImageNet\cite{deng2009imagenet} which contains 100,000 images of 200 classes. Each class has 500 training images, 50 validation images, and 50 test images. (3) ImageNet \cite{deng2009imagenet} dataset, which contains 1,200,000 images for training and 50,000 images for validation over 1,000 classes.

\subsubsection{Implementation Details}
\label{sec:implement}
To prove the effectiveness of CAKD, we adopt Resnet \cite{he2016deep}, WideResNet(WRN) \cite{zagoruyko2016wide}, ShuffleNet \cite{zhang2018shufflenet,ma2018shufflenet} and MobileNet \cite{sandler2018mobilenetv2} as the backbone network. We follow the training setting of \cite{tian2019crd, zhao2022decoupled}, except for the batch size and the method for Multi-GPU training.

We train the teacher model and the student model for 240 epochs with SGD. As the batch size is 128, the learning rates are 0.05 for ResNet\cite{he2016deep} and WRN\cite{zagoruyko2016wide} and 0.01 for ShuffleNet \cite{zhang2018shufflenet,ma2018shufflenet}. The learning rate is divided by 10 at 150, 180, and 210 epochs. The weight decay and the momentum are set to 5e-4 and 0.9, and the temperature is set to 4. We utilize a warm-up of 20 epochs for all experiments as well. The weight for the hard label loss is set to 1.0. The weights for BCD, SCD, and WCD on the feature level and BCD and WCD on the logit level are different for different experiments.

\subsection{Performances}
\label{sec:result}
\subsubsection{Comparison on the CIFAR-100}
\label{sec:cifar}
Tab.\ref{tab:cifar100} shows the results of student models on the CIFAR-100 with different teacher-student architectures, which can be grouped into same architecture style and different architecture style. For each teacher-student architecture pair, we use the same pre-trained teacher model for a fair comparison. Overall, our CAKD outperforms other baseline methods.

\subsubsection{Comparison on the Tiny-ImageNet}
\label{sec:tiny}
Tab.\ref{tab:tiny} shows the results of student models on the Tiny-ImageNet with different teacher-student architectures. Same to Tab.\ref{tab:cifar100}, the results can be grouped into the same architecture style and different architecture styles. Compared to the performance on the CIFAR-100, the improvements on the Tiny-Imagenet are smaller. The reason is the clustering for $\mathbf{S}$ and $\mathbf{W}$ relies on the confidence of the teacher model. Inaccurate clustering will increase the difficulty of generating the appropriate feature mask. 

\subsubsection{Comparison on the ImageNet}
\label{sec:imagenet}
Tab.\ref{tab:imagenet} shows the results of student models on the ImageNet with different teacher-student architectures. For both same architecture style and different architecture style, our CAKD also achieves better performance than other baselines, which proves our approach works well on large-scale datasets.

\subsection{Ablation Studies}
\label{sec:analysis}

Given the hyperparameters at the logit level have been thoroughly explored in \cite{zhao2022decoupled}, we follow its hyperparameters settings at the logit level and concentrate on the feature level in the following ablation studies.

\subsubsection{Effect of BCD, SCD \& WCD}
 
In Eqn.\eqref{eq:7}, BCD represents the accuracy of our distinction between the strong correlation cluster $\mathbf{S}$ and the weak correlation cluster $\mathbf{W}$. SCD represents the divergence of feature values within the $\mathbf{S}$, while WCD represents the divergence of feature values within the $\mathbf{W}$. As can be inferred Eqn.\eqref{eq:7}, compared to the unaffected BCD, SCD and WCD are suppressed by $p_s^{\mathcal{T}}$ and $p_w^{\mathcal{T}}$, respectively. To achieve better accuracy, we should reinforce the effects of SCD and WCD, which means increasing the values of $\alpha$ and $\beta$. Among SCD and WCD, SCD usually plays a more crucial role. However, it's worth noting that the distinction between clusters $\mathbf{S}$ and $\mathbf{W}$ heavily relies on the teacher model's confidence. Therefore, the relationship between $\alpha$ and $\beta$ should also consider the teacher model's confidence.  Intuitively, the more confident the teacher is, the more valuable the WCD could be, and the smaller $\alpha/\beta$ should be applied. Thus, we suppose that the value of $\alpha/\beta$ could be related to the feature value of $SCD$ and $WCD$. Specifically, the relationship between the summarization of $SCD$ and $WCD$ can be denoted as:

\begin{equation}
\begin{aligned}
\label{eq:11}
    p_s/p_w &= \frac{\sum_{i=1, i \in S}^C \exp \left(f_i\right)}{\sum_{k=1}^C \exp \left(f_k\right)} / \frac{\sum_{i=1, i \notin S}^C \exp \left(f_i\right)}{\sum_{k=1}^C \exp \left(f_k\right)}\\
    &= \frac{\sum_{i=1, i \in S}^C \exp \left(f_i\right)}{\sum_{i=1, i \notin S}^C \exp \left(f_i\right)}
\end{aligned}
\end{equation}

\begin{table}[t]
\caption{Top-1 accuracies(100\%) on the ImageNet.} 
\begin{center}
\begin{tabular}{c|cc}
  \hline
    & Same Architecture & Different Architecture \\
  \hline
  Teacher  & ResNet34  & ResNet50  \\
  Acc   &   73.31      &  76.16  \\
  Student & ResNet18 & MobileNetV2\\
  Acc &  69.75    &  68.87  \\
  \hline
  KD\cite{hinton2015distilling} & 70.66 &  68.58 \\
  CRD\cite{tian2019contrastive} & 71.17 &  71.37 \\
  OFD\cite{heo2019comprehensive} & 70.81 &  71.25 \\
  CTKD\cite{li2023curriculum} & 71.09 & 71.11\\
  ReviewKD\cite{chen2021distilling} & 71.61 & 72.56 \\
  DKD\cite{zhao2022decoupled} & 71.70 & 72.05 \\
  NKD\cite{yang2023knowledge} & 72.13 & 72.96\\
  \hline
  CAKD & \textbf{72.17}  & \textbf{73.41} \\
  \hline
\end{tabular}
\label{tab:imagenet}
\end{center}
\end{table}

To elucidate the aforementioned perspectives and to inform hyper-parameter optimization, we showcase the effects of different $\alpha$ and $\beta$ on the student model's accuracy in Tab.\ref{tab:effect_c} and Tab.\ref{tab:effect_t}. We can observe that increasing $\alpha$ and $\beta$ to appropriate values can enhance the accuracy of the student model. For the CIFAR-100 dataset, which boasts a higher teacher accuracy, the impact of $\alpha$ generally outweighs that of $\beta$. On the other hand, for the Tiny-ImageNet dataset with relatively lower accuracy, the value of $\alpha/\beta$ needs to be lower to achieve the highest accuracy, which means we should not suppress the effect of WCD overly.

\subsubsection{Balancing Multiple Distillation Components}
CAKD loss consists of multiple weighted distillation components. To investigate the effect of different distillation components, we use different $\gamma$ to adjust the proportionality between the feature distillation component and logit distillation component. Cause the value of the loss is highly related to the size of the feature or the logit, we present our experiments based on the scaling relationship of the loss rather than the scaling relationship of the $\gamma$ in Tab.\ref{tab:terms}. 

The ResNet is conducted on CIFAR-100 and the WRN is conducted on Tiny-ImageNet. We can find all experiments show that feature loss and logit loss should play a similar role to achieve the best performance. If we take a further comparison of these accuracies, we can see that the accuracies based on Tiny-ImageNet are more sensitive to the adjustment of the loss than CIFAR-100, when the loss of feature plays a much more important role than the loss of logit, specifically. This may be due to the lower teacher model confidence which makes the mask generated by the scoring network more tricky.

\subsubsection{Effect of More Distillation Components}
Ablation experiments of multiple distillation components are conducted to measure their effects. The result is shown in Tab.\ref{tab:layers}, the ResNet is validated on CIFAR-100 and WRN is validated on Tiny-ImageNet. From the table, we can observe that distilling either logit or feature individually can achieve positive results, distilling more layers tends to yield better performance. 


\begin{table}[t]
\caption{Top-1 accuracies(100\%) on the CIFAR-100 with different $\alpha$ and $\beta$.} 
\begin{center}
\begin{tabular}{cc|ccc}
  \hline
  \multicolumn{2}{c|}{Teacher}  & ResNet56 & ResNet32x4& WRN40-2\\
  \multicolumn{2}{c|}{Acc}  &   72.32   & 79.32 & 75.50\\
  \multicolumn{2}{c|}{$p_s/p_w$}  &  3.37   & 3.45 & 3.13\\
  \hline
  &2,2 & 71.69 & 75.68 & 75.64\\
 & 4,4&  71.69 & 75.67 & 75.73 \\
 $\alpha ,\beta =$& 8,4&  71.86 & 75.89 & 75.84 \\
 & 8,2&  \textbf{72.17} & \textbf{75.99} & \textbf{75.95} \\
 & 8,1&  71.72 & 75.76 & 75.69 \\
  \hline

\end{tabular}
\label{tab:effect_c}
\end{center}
\end{table}

\begin{table}
\caption{Top-1 accuracies(100\%) on the Tiny-ImageNet with different $\alpha$ and $\beta$.} 
\begin{center}
\begin{tabular}{cc|ccc}
  \hline
  \multicolumn{2}{c|}{Teacher}  & ResNet56 & ResNet32x4& WRN40-2\\
  \multicolumn{2}{c|}{Acc}  &   50.78   & 60.17 & 56.06\\
   \multicolumn{2}{c|}{$p_s/p_w$}  &  2.51   & 2.24 & 2.39\\
  \hline
 &2,2& 44.40 &50.84 &51.15 \\
 & 4,4&44.60 &51.23 &51.22\\
$\alpha ,\beta =$ & 8,4&\textbf{44.85} &\textbf{51.59} &\textbf{51.40}\\
 & 8,2&44.77 &51.54 &51.33\\
 & 8,1&44.75 &51.39 &51.37\\
  \hline

\end{tabular}
\label{tab:effect_t}
\end{center}
\end{table}

\begin{table}[t]
\caption{Top-1 accuracies(100\%) with different loss proportionality among distillation components.} 
\begin{center}
\begin{tabular}{cc|cc}
  \hline
  \multicolumn{2}{c|}{Teacher}  & ResNet56-1 & ResNet56-2 \\
  \multicolumn{2}{c|}{Acc}  &   72.32       &  50.78    \\
  \hline
 &0.2 &71.47 &44.28\\
 & 0.5& 71.69&44.61\\
 $L_{feature} / L_{logit}=$& 1& \textbf{71.27}&\textbf{44.93}\\
 & 2&71.45 &44.19\\
 & 5&70.89 &43.34\\
  \hline

\end{tabular}
\label{tab:terms}
\end{center}
\end{table}

\begin{table}
\caption{Top-1 accuracies(100\%) with different distillation components.} 
\begin{center}
\begin{tabular}{c|cc}
  \hline
  Teacher  & ResNet56  & WRN40-2  \\
  Acc   &   72.32       &  56.06  \\
  \hline
  logit & 71.69 &   50.36\\
  single feature & 71.15 & 49.72\\
  logit \& single feature & 72.27 & 50.97\\
  logit \& double features &\textbf{72.31} & \textbf{51.09}\\
  \hline

\end{tabular}
\label{tab:layers}
\end{center}
\end{table}

\section{Conclusion and Discussion}
\label{sec:discussion}



In this study, we have proposed an innovative approach that decouples KL divergence, aiming to scrutinize the impact of each element in the distillation process. We have demonstrated that by striking a balance between the enhanced SCD and WCD, we can amplify the influence of elements that have a significant impact on predictions.In our future work, we intend to extend our investigation to include multi-class classification scenarios, thereby broadening the application spectrum for our CAKD framework.

\end{document}